\pdfoutput=1
\documentclass[letterpaper, preprint, paper,11pt]{AAS}	% for preprint proceedings

\usepackage{siunitx}                                                         
\usepackage{hyperref}                         \newcommand{\R}{\mathbb{R}}                      
\hypersetup{    colorlinks=true,
                linktoc=all,
                linkcolor=black,
                citecolor=black,}
\hypersetup{linktocpage}

\usepackage{mathtools}                 % Load standard math package
\usepackage{cases}                  % Provides improved piecewise env.
\usepackage{amssymb}            % Load AMS symbol package (for mathbb{})
\usepackage[makeroom]{cancel}
\usepackage{gensymb}                   % Load generic symbol package
\usepackage{wasysym}                    % Load even fancier symbol package
\usepackage{bm}                    % Better than mathbf{}
\usepackage{arydshln}               % Allows for dashed matrix separators
\usepackage{float}                                              % Enables float environment
\usepackage{wrapfig}                    % Wrap figures/tables in text (i.e., Di Vinci style)
\usepackage{threeparttable}                  % Tables with footnotes
\usepackage{multirow}                                                 
\usepackage{hhline}
\usepackage{tikz}
\usetikzlibrary{shapes,arrows,positioning}
\usepackage[font=bf]{caption}          % Caption customization
\usepackage[section]{placeins}                     % Enables \FloatBarrier command

\usepackage[nameinlink]{cleveref}      % Enables extended referencing
\crefname{figure}{Figure}{Figures}
\Crefname{figure}{Figure}{Figures}
\crefname{equation}{Eq.}{Eqs.}
\Crefname{equation}{Equation}{Equations}
\crefname{table}{Table}{Tables}
\Crefname{table}{Table}{Tables}
\crefname{appsec}{Appendix}{Appendices} % Doesn't work with \section*
\Crefname{appsec}{Appendix}{Appendices} % Doesn't work with \section*

\title{Modeling of an On-Orbit Maintenance Robotic Arm Test-Bed}

\author{Jacob J. Korczyk\thanks{Systems Engineer, Raytheon Technologies, 22260 Pacific Blvd, Sterling, VA 20166}, Daniel Posada\thanks{PhD Student, Aerospace Engineering, Embry–Riddle Aeronautical University 600 S Clyde Morris Blvd, Daytona Beach, FL 32114}, Aryslan Malik \thanks{PhD Candidate, Aerospace Engineering, Embry–Riddle Aeronautical University 600 S Clyde Morris Blvd, Daytona Beach, FL 32114 }, and Troy Henderson\thanks{Associate Professor, Aerospace Engineering, Embry–Riddle Aeronautical University 600 S Clyde Morris Blvd, Daytona Beach, FL 32114}}

\PaperNumber{21-760}

\begin{document}

\maketitle

%%%%%%%%%%%%%%%%%%%%%%%%%%%%%%%%%%%%%%%%%%%%%%%%%%%%%%%%%%%%%%%%%%%%%%
\begin{abstract}
\indent This paper focuses on the development of a ground-based test-bed to analyze the complexities of contact dynamics between multibody systems in space.
The test-bed consists of an air-bearing platform equipped with a 7 degrees-of-freedom (one degree per revolute joint) robotic arm which acts as the servicing satellite.
The dynamics of the manipulator on the platform is modeled as an aid for the analysis and design of stabilizing control algorithms suited for autonomous on-orbit servicing missions.

The dynamics are represented analytically using a recursive Newton-Euler multibody method with D-H parameters derived from the physical properties of the arm and platform. In addition, Product of Exponential (PoE) method is also employed to serve as a comparison with the D-H parameters approach. Finally, an independent numerical simulation created with the SimScape\textsuperscript{\tiny{\textrm{TM}}} modeling environment is also presented as a means of verifying the accuracy of the recursive model and the PoE approach.
The results from both models and SimScape\textsuperscript{\tiny{\textrm{TM}}} are then validated through comparison with internal measurement data taken from the robotic arm itself.
\end{abstract}

%%%%%%%%%%%%%%%%%%%%%%%%%%%%%%%%%%%%%%%%%%%%%%%%%%%%%%%%%%%%%%%%%%%%%%
\section{Introduction}
As humans endeavor to conduct missions of increasing complexity, both in Earth orbit and beyond, operations such as asteroid mining, space debris removal, and on-orbit servicing will have to be performed routinely. To model the complexities of the contact dynamics involved in these procedures a robotics test-bed design is proposed consisting of two interacting robotic arms. The first robotic arm is mounted on an air-bearing platform, acting as the servicing satellite, while the second arm is stationary, acting as the client spacecraft.

This test-bed environment is essential to continue the innovation process and improve space exploration, as there has been an increased interest in servicing missions.
By testing here on earth, the dangers of on-orbit servicing can be evaluated and assessed.
Different examples of robotic systems in space can be found in literature \cite{currie2002international}, three of them reside on the International Space Station (ISS) such as: The Canadian Mobile Servicing System (MSS), which is used to perform maintenance to the station and provide aid for approaching maneuvers,
the Japanese (JAXA) Experiment Module Robotics System (JEMRMS) which is used for experimentation and payload maneuvers, and finally, the European Space Agency (ESA) European Robotic Arm (ERA), used for maintenance and extravehicular activity support.
These examples are all implemented on a system with considerably more mass and volume than the manipulator. 
However, with the current trend in miniaturization, smaller, more agile, spacecraft will need to carry out similar missions, like what has been done by Antonello \cite{Dynamics2018} using thrusters and Schwartz \cite{schwartz2004} for decentralized control. 

\begin{figure}
    \centering
    \includegraphics[width=\textwidth]{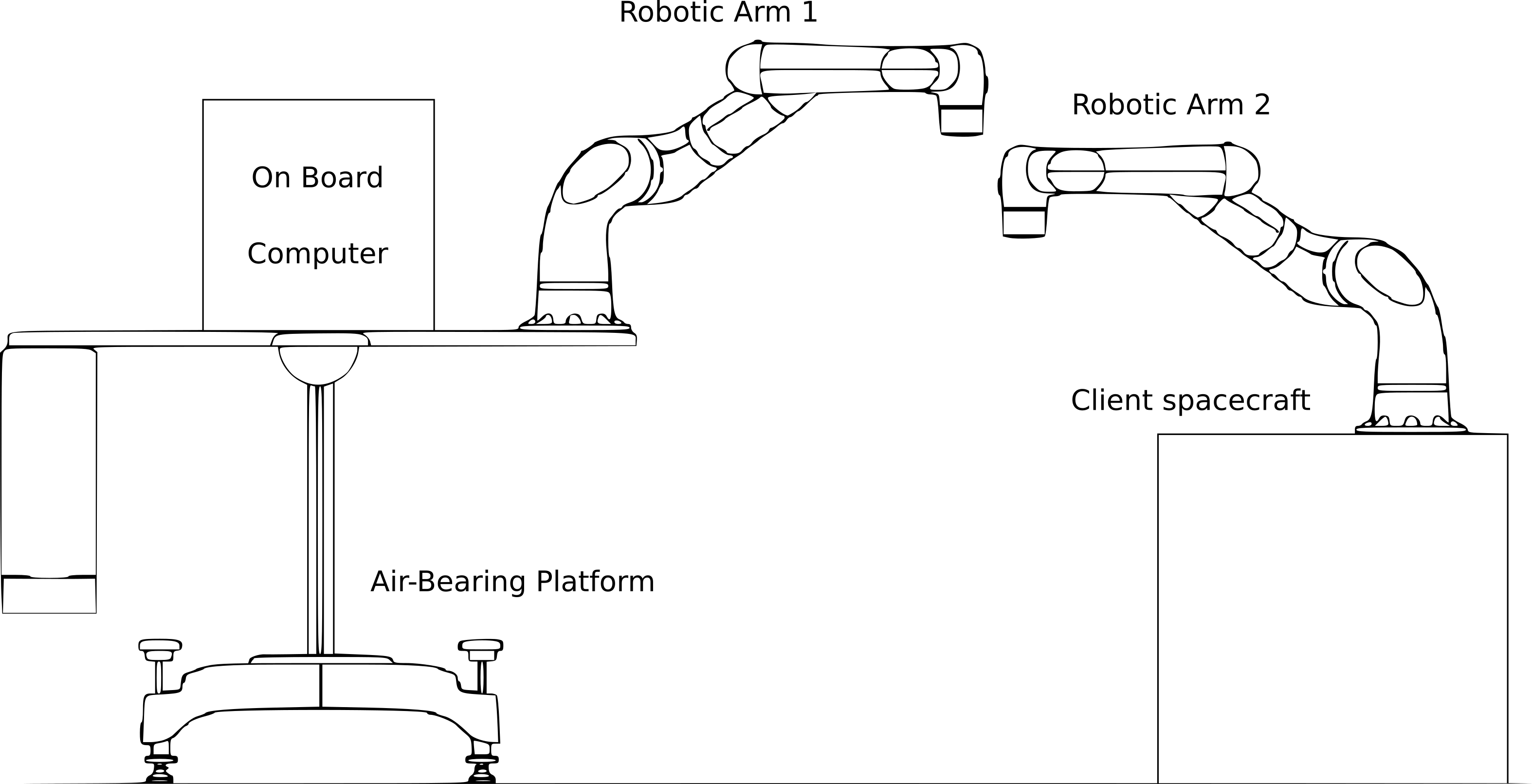}
    \caption{Space Robot System}
    \label{fig:robot_arms}
\end{figure}

Because of the small size of this satellite when the robotic arm is actuated in orbit, as there are no external influences, the attitude of the satellite will change as a consequence of this motion.
This effect is undesirable as it can cause unwanted maneuvers that could have drastic effects on communication, power generation, even compromise the mission.
Using the air-bearing test-bed these dynamics can be studied, and their effects better understood, giving the possibility to develop controllers rejecting said undesired phenomena using different control approaches.

Each robotic arm on the air-bearing test-bed features seven single degree of freedom revolute joints. When mounted on the air bearing platform, the assembly gains an additional three degrees of freedom due to the spherical air bearing acting as a universal joint. Because of its kinematic complexity, a recursive method is used to model the system. To generate the equations of motion the kinematic properties are propagated in a forward recursion from the first to the last link to generate velocities and accelerations. The process is then conducted in reverse, from the last to the first link to determine the forces and torques at each joint.

For validation purposes, a numerical simulation was created through the use of MATLAB\textsuperscript{\textregistered} and the SimScape\textsuperscript{\textrm{TM}} Multibody environment \cite{SimScapeManual}. Torque results generated by both the recursive (D-H and PoE) and numerical models will be compared. This model is now used in both the synthesis and analysis of control algorithms for the simulator as well as a design aid for the platform itself. The implementation of these control algorithms are outlined in Korczyk's work \cite{korczyk2020dynamic}.

%%%%%%%%%%%%%%%%%%%%%%%%%%%%%%%%%%%%%%%%%%%%%%%%%%%%%%%%%%%%%%%%%%%%
\section{Robot arm Dynamics}
To analyze the dynamics of the robotic arm, the mathematical model formulation described by Ploen \cite{Ploen97} which was developed originally by Park \cite{Park94}, was used; yielding the following recursive algorithm for an open chain serial manipulator:
\begin{equation}\label{Torque}
    \tau = M(q)\ddot{q} + C(q,\dot{q})\dot{q} + \phi(q)
\end{equation}
where $M(q)$ denotes the mass matrix, $C(q,\dot{q})$ is the Coriolis/centrifugal matrix, $\phi(q)$ denotes the gravity terms, and $\tau$ represents the the applied torques at each joint:
\begin{equation}
    M(q) = S^TG^TJGS
\end{equation}
\begin{equation}
    ad_{S_{\dot{q_i}}} = \left[\begin{matrix}-[(S_{\dot{q_i}})^\times] & 0 \\ 0 & -[(S_{\dot{q_i}})^\times]
    \end{matrix} \right]
\end{equation}
\begin{equation}
    ad_{V_i}^*=\begin{bmatrix}
        -[\Vec{\omega}_{i}^\times]&-[\Vec{v}_{i}^\times]\\
           0 & -[\Vec{\omega}_{i}^\times] \\
         \end{bmatrix}
\end{equation}
\begin{equation}
    C(q,\dot{q}) = S^TG^T(JGad_{S_{\dot{q}}}\Gamma+ad_{V}^{*}J)GS
\end{equation}
\begin{equation}
    \phi(q) = S^TG^TJGP_0\dot{V}_0
\end{equation}
and the definition for each term is:
$$ad_{S_{\dot{q}}}=diag[ad_{S_{\dot{q_1}}},ad_{S_{\dot{q_2}}},\dots,ad_{S_{\dot{q_n}}}]$$
$$ad_{V}^*=diag[ad_{V_1}^*,ad_{V_2}^*,\dots,ad_{V_n}^*]$$
$$\dot{q}=[\dot{q}_1,\dot{q}_2,\dots,\dot{q}_n]^T$$
$$\tau=[\tau_1,\tau_2,....,\tau_n]^T$$
$$P_0 = [Ad_{f_{0,1}^{-1}},0_{6\times6},\dots,0_{6\times6}]^T$$
$$S = diag[s_1,s_2,\dots,s_n]$$
$$J = diag\left[\left[\begin{matrix}I_1-m_1[r_1]^2 & m_1[r_1]\\-m_1[r_1] & m_1I_{3\times3}\end{matrix}\right],\dots,\left[\begin{matrix}I_n-m_n[r_n]^2 & m_n[r_n]\\-m_n[r_n] & m_nI_{3\times3}\end{matrix}\right]\right]$$
$$G = (1-\Gamma)^{-1} = I + \Gamma + \dots + \Gamma^{n-1}$$
$$\Gamma = \left[\begin{matrix}0_{6\times6} & 0_{6\times6} & \dots & 0_{6\times6} & 0_{6\times6}\\ Ad_{f_{1,2}^{-1}} & 0_{6\times6} & \dots & 0_{6\times6} & 0_{6\times6}\\ 0_{6\times6} & Ad_{f_{2,3}^{-1}} & \dots & 0_{6\times6} & 0_{6\times6}\\ \vdots & \vdots & \ddots & \dots & \dots\\
0_{6\times6} & 0_{6\times6} & \dots & Ad_{f_{n-1,n}^{-1}} & 0_{6\times6}\end{matrix}\right]$$
$$
\dot{V}_0=[0,0,0,0,0,g]^T$$
where, $S$ matrix is
composed of smaller vectors, $s_i$
, that are composed of the zero vector and the unit
vector along the axis of rotation\cite{korczyk2020dynamic}:
\begin{equation}\label{eq9}
    {s}_i = \begin{bmatrix}
        \Vec{\omega}_i\\
           0_{1\times3} \\
         \end{bmatrix}\in\R^6
\end{equation}
This method was developed using Lie group algebra formulation as the dynamics of the arm can be differentiated in a straightforward manner. Because the method is solely based on the matrix exponential, a basic mathematical primitive \cite{Ploen95}.

The Product of Exponentials - Forward Kinematics (PoE-FK) formulation is presented as follows \cite{malik2021trajectory}:
\begin{equation}\label{eq7}
      {T_{sn}}=e^{[\mathcal{S}_1]\theta_1}e^{[\mathcal{S}_2]\theta_2}\cdots e^{[\mathcal{S}_n]\theta_n}M_{sn}
\end{equation}
where $T_{sn}\in{SE}(3)$ is the configuration of an $n$-th point on the manipulator; $M_{sn}\in{SE}(3)$ is the $n$-th point's home configuration; $\mathcal{S}_i\in\R^6$ is a ``unit" screw axis represented in the inertial (space) frame such that $\left\lVert \omega\right\rVert=1$ and $v$ is the linear velocity at the inertial-frame origin, expressed in the inertial frame produced purely due to the rotation about the $i$-th screw axis $(v=-\omega\times{r_{q}})$, where $q$ is a point on the screw axis:
\begin{equation}\label{eq9}
    \mathcal{S}_i = \begin{bmatrix}
           \Vec{\omega}_i \\
           \Vec{v}_i \\
         \end{bmatrix}\in\R^6
\end{equation}
Screw axes are defined by the rolling and pitching joints of the Sawyer robotic manipulator, which makes the forward kinematics simpler in implementation, since, compared to D-H parameters, PoE does not require initializing frames for each joint. The screw axes are defined by the physical locations of the joints, for Sawyer robotic manipulator the screw axes are given as follows:
\begin{align}\label{eq:screw}
    \mathcal{S}_1 &=\begin{bmatrix}
0 \\
0 \\
1 \\
0 \\
0 \\
0 \\
         \end{bmatrix}\;\;\;\;\;\;\;\;\;\;\;\;\;\;
    \mathcal{S}_2 = \begin{bmatrix}
0         \\
1         \\
0         \\
-0.317    \\
0         \\
0.081     \\
         \end{bmatrix}\;\;\;\;\;
    \mathcal{S}_3 = \begin{bmatrix}
1        \\
0        \\
0        \\
0        \\
0.317    \\
-0.1925  \\
         \end{bmatrix}\;\;\;\;\;
    \mathcal{S}_4 = \begin{bmatrix}
0          \\
1          \\
0          \\
-0.317     \\
0          \\
0.481      \\
         \end{bmatrix}\\
    \mathcal{S}_5 &= \begin{bmatrix}
1       \\
0       \\
0       \\
0       \\
0.317   \\
-0.024  \\
         \end{bmatrix}\;\;\;\;\;
    \mathcal{S}_6 = \begin{bmatrix}
0           \\
1           \\
0           \\
-0.317      \\
0           \\
0.881       \\
         \end{bmatrix}\;\;\;\;\;
    \mathcal{S}_7 = \begin{bmatrix}
1         \\
0         \\
0         \\
0         \\
0.317     \\
-0.1603   \\
         \end{bmatrix}
  \end{align}
Square brackets around screw axes represent skew-symmetric mapping such that: $\Vec{p}\in\R^3\,	{\rightarrow}\,{[\Vec{p}]}\in{so(3)}$ and ${\mathcal{S}}\in\R^6\,	{\rightarrow}\,{[\mathcal{S}]}\in{se(3)}$. For instance:
\begin{align}\label{skew-sym}
    [\mathcal{S}]=\begin{bmatrix}
           [\Vec{\omega}] & \Vec{v} \\
           0 & 0 \\
         \end{bmatrix}
\end{align}
PoE-FK utilizes exponential mapping such that: ${[\mathcal{S}]}\theta\in{se(3)}\,	{\rightarrow}\,T\in{SE(3)}$.

The PoE inverse dynamics algorithm is carried out by employing the following formulation \cite{malik2021trajectory}:

\begin{equation}\label{eq:torque_poe}
       {\tau} = M({\theta})\Ddot{\theta} + h({\theta},{\dot{\theta}})
\end{equation}
where $\theta\in{\R}^n$ is the vector of joint variables, $\tau\in{\R}^n$ is the vector of joint torques, $M({\theta})\in{\R}^{n{\times}n}$ is the symmetric positive-definite configuration-dependent mass matrix, and $h({\theta},{\dot{\theta}})\in{\R}^n$ is the combination of centripetal, Coriolis, gravity, and friction terms that depend on $\theta$ and $\dot{\theta}$. Exhaustive derivation process of $M({\theta})$ and $h({\theta},{\dot{\theta}})$ matrices is outlined in Malik's work \cite{malik2021trajectory}.
%%%%%%%%%%%%%%%%%%%%%%%%%%%%%%%%%%%%%%%%%%%%%%%%%%%%%%%%%%%%%%%%%%%%
\section{MATLAB Simulation of Robotic Arm}

\begin{figure}[h!]
\begin{minipage}{.45\textwidth} %
    \centering
    \includegraphics[scale=0.9]{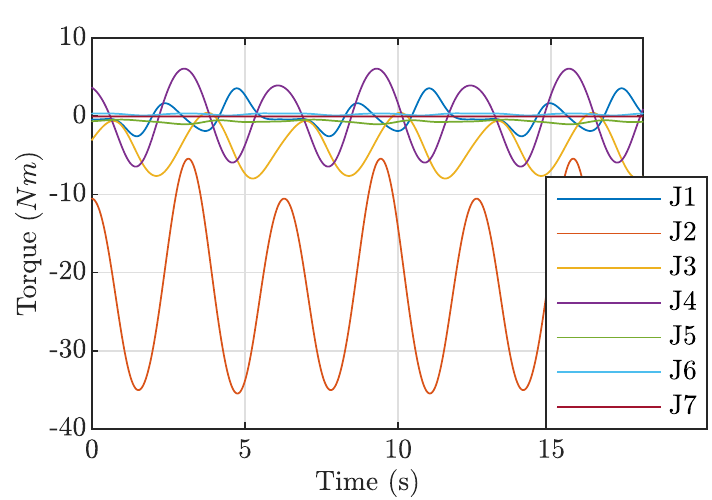}
    \caption{Joint torques generated by D-H parameters method}
    \label{fig:dhsin}
\end{minipage} %
\begin{minipage}{.45\textwidth} %
    \centering
    \includegraphics[scale=0.9]{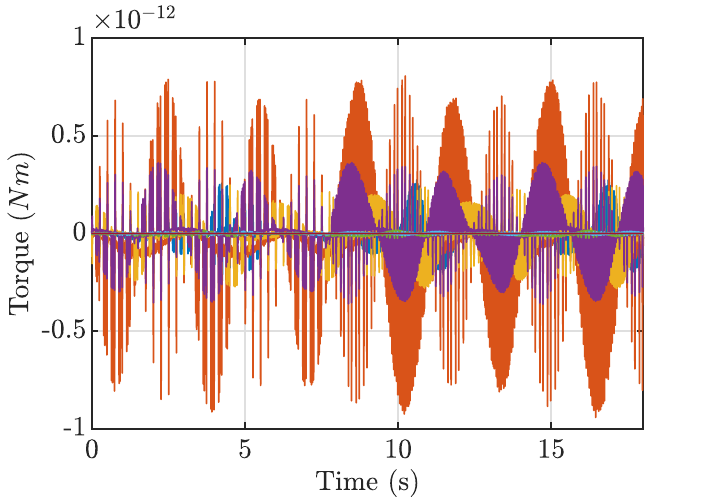}
    \caption{Relative error propagation between D-H and SimScape\textsuperscript{\textrm{TM}}}
    \label{fig:dhsinerror}
\end{minipage} %
\end{figure}

\begin{figure}[h!]
\begin{minipage}{.45\textwidth} %
    \centering
    \includegraphics[scale=0.9]{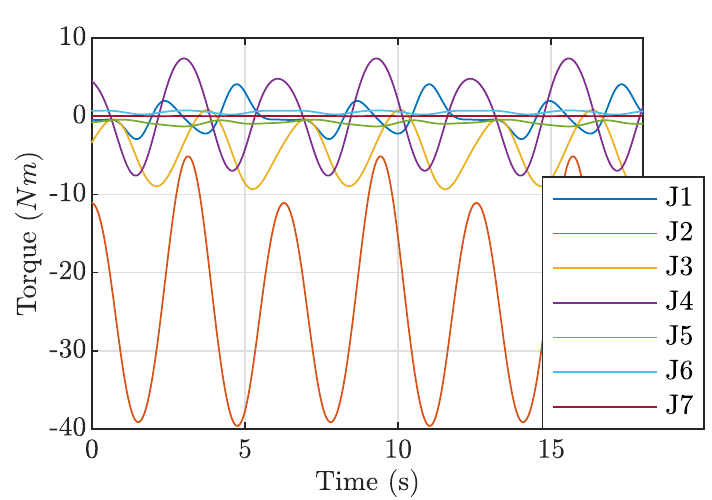}
    \caption{Joint torques generated by PoE method}
    \label{fig:poesin}
\end{minipage} %
\begin{minipage}{.45\textwidth} %
    \centering
    \includegraphics[scale=0.9]{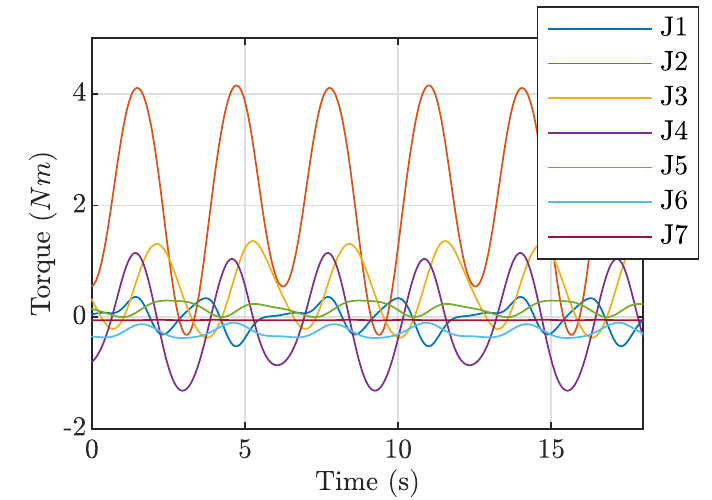}
    \caption{Relative error propagation between PoE and SimScape\textsuperscript{\textrm{TM}}}
    \label{fig:poesinerror}
\end{minipage} %
\end{figure}

Using this recursive method the torques for each joint can be calculated using D-H paramters approach shown in \cref{Torque} or PoE approach described by \cref{eq:torque_poe}. 
To generate the torque and the angular positions, velocities and accelerations of each joint must be specified. 
The specific trajectories are provided as functions of time as seen in \cref{tab:PVA}, that result in the torques plotted in \cref{fig:dhsin} and \cref{fig:poesin}.
Modeled in SimScape\textsuperscript{\textrm{TM}}, the numerical simulation of the dynamics described analytically by the D-H parameters model generates identical results as shown by the subtraction of the results of the D-H parameters method from that of the numerical simulation (\cref{fig:dhsinerror}). The PoE method, on the other hand, shows different torque profile as demonstrated in \cref{fig:poesinerror}, which indicates that the SimScape\textsuperscript{\textrm{TM}} and D-H approaches calculate the torque profiles in a different way.

This was performed to analyze the veracity of both mathematical approaches, D-H and PoE methods performed completely coded using functions developed by the team and exclusively from the code editor in MATLAB. The SimScape\textsuperscript{\textrm{TM}} numerical method described the algorithm graphically using the Simulink environment through a modular block approach.

\begin{table}[h!] 
    \centering
\begin{tabular}{|c|c|c|c|}
\hline\noalign{\vskip-2.5pt}
\rule{0pt}{14pt}
 Joint & Position & Velocity & Acceleration  \\
\hhline{|=|=|=|=|}
   1 & sin(t) & cos(t) & -sin(t) \\ 
\hline
   2  & cos(t) & -sin(t) & -cos(t) \\
\hline
   3 & sin(t) & cos(t) & -sin(t) \\ 
\hline
   4 & cos(t) & -sin(t) & -cos(t) \\
\hline
   5 & sin(t) & cos(t) & -sin(t) \\ 
\hline
   6 & cos(t) & -sin(t) & -cos(t) \\
\hline
   7 & sin(t) & cos(t) & -sin(t)  \\ 
\hline
\end{tabular}
    \caption{Joint Trajectories}
    \label{tab:PVA}
\end{table}


Joint torque profiles generated using recursive method (Figure \ref{fig:dhsin}) and PoE inverse dynamics method (Figure \ref{fig:poesin}) produce different torques throughout the trajectory defined by \cref{tab:PVA}. Joint 2 shows the most discrepancy where the difference between PoE and D-H torque trajectories reach maximum value of $4\;Nm$. This could be caused by the assumptions done when modeling the links using D-H parameters when typing the code and modeling in Simulink. More information on these discrepancies will be detailed in the following sections.

%%%%%%%%%%%%%%%%%%%%%%%%%%%%%%%%%%%%%%%%%%%%%%%%%%%%%%%%%%%%%%%%%
\section{Validation of the Model}
\begin{figure}[h!]
\begin{minipage}{.45\textwidth} %
    \centering
    \includegraphics[scale=0.8]{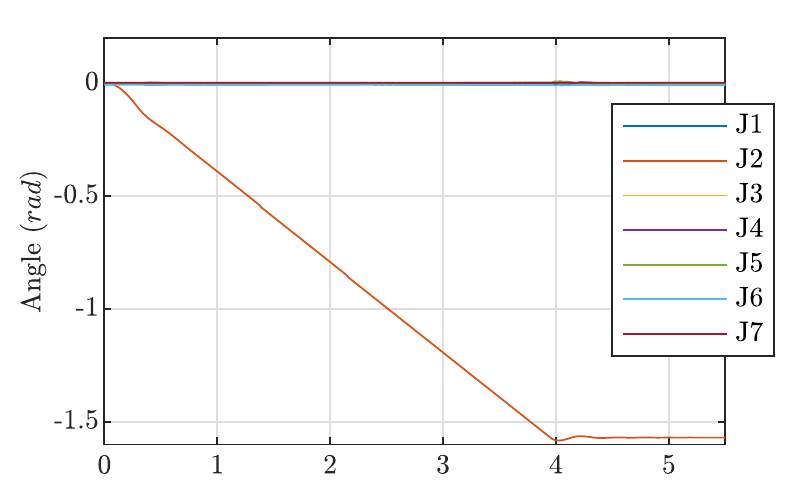}
    \caption{Experimental joint angles}
    \label{fig:exp_angles}
\end{minipage} %
\begin{minipage}{.45\textwidth} %
    \centering
    \includegraphics[scale=0.8]{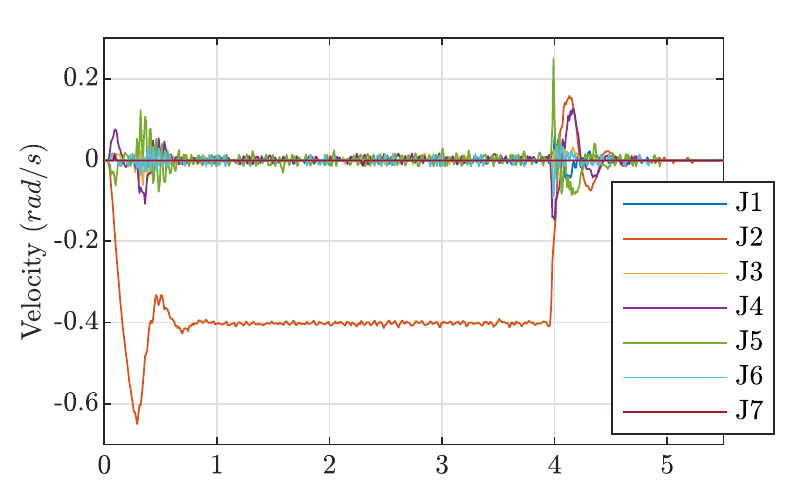}
    \caption{Experimental joint velocities}
    \label{fig:exp_vel}
\end{minipage} %
\end{figure}

\begin{figure}[h!]
\begin{minipage}{.45\textwidth} %
    \centering
    \includegraphics[scale=0.8]{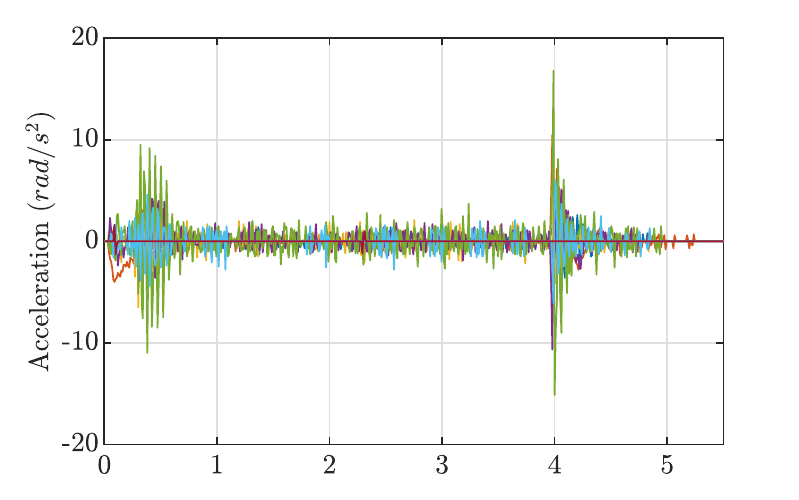}
    \caption{Experimental joint accelerations}
    \label{fig:exp_acc}
\end{minipage} %
\begin{minipage}{.45\textwidth} %
    \centering
    \includegraphics[scale=0.8]{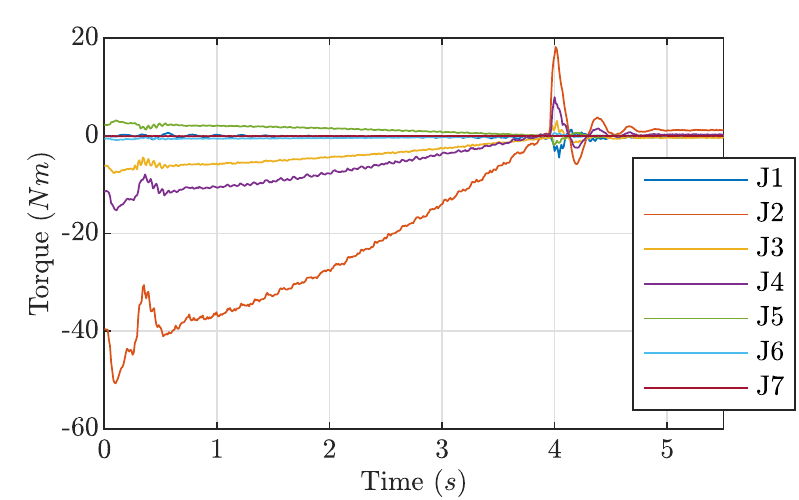}
    \caption{Experimental joint torques}
    \label{fig:exp_torq}
\end{minipage} %
\end{figure}

\begin{figure}[h]
\begin{minipage}{.45\textwidth} %
    \centering
    \includegraphics[scale=0.8]{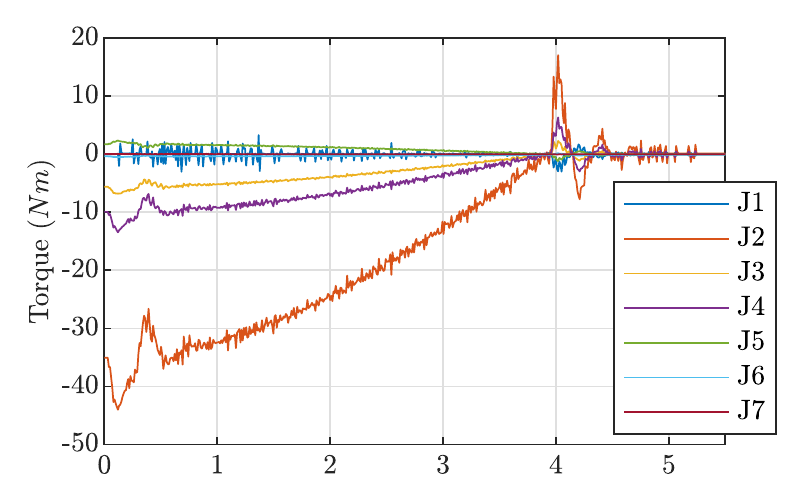}
    \caption{Simulated torque profile generated by D-H}
    \label{fig:dh_torq}
\end{minipage} %
\begin{minipage}{.45\textwidth} %
    \centering
    \includegraphics[scale=0.8]{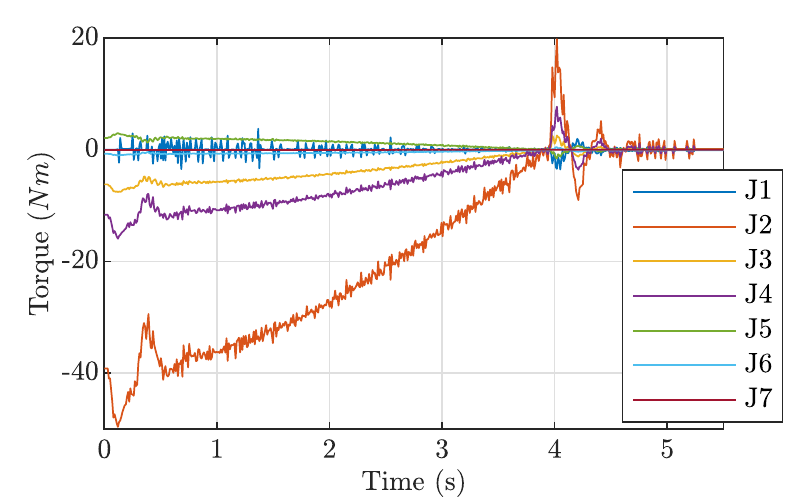}
    \caption{Simulated torque profile generated by PoE}
    \label{fig:poe_torq}
\end{minipage} %
\end{figure}

To ensure the validity of D-H parameters approach and PoE approach an experiment was conducted to compare the simulated torque trajectories with experimental ones. The comparison proved the accuracy of both methods, and it was observed that the PoE approach is more accurate than the D-H parameters approach. 
The trajectory for these tests was generated by commanding the robot to move from its zero position, all links were configured to extend to the robot's furthest reach in the $x$-direction (this is defined as the rest position). Afterward, all links were configured to extend to the robot's furthest reach directly to the upward pose in the $z$-direction. The joint positions, velocity, acceleration, and torque profiles corresponding to this motion are shown in Figures \ref{fig:exp_angles}-\ref{fig:exp_torq}. The simulated torque profiles of D-H and PoE approaches are shown in Figures \ref{fig:dh_torq} and \ref{fig:poe_torq} respectively.


\begin{figure}[h]
\begin{minipage}{.45\textwidth} %
    \centering
    \includegraphics[scale=0.8]{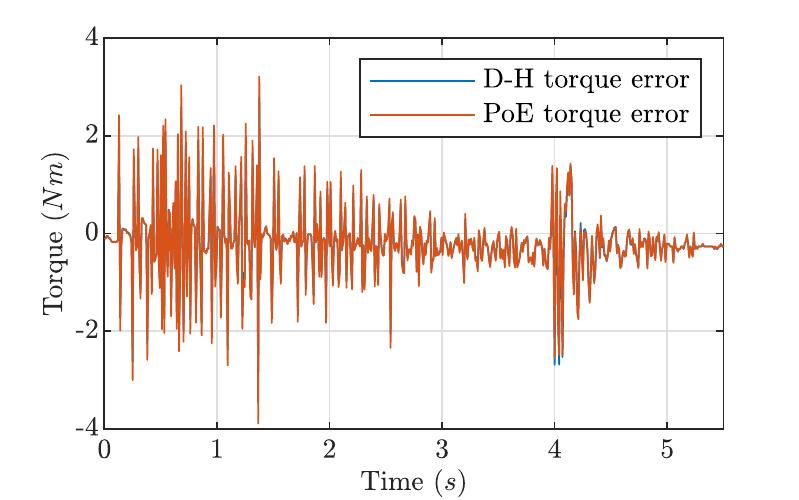}
    \caption{Joint 1 error}
    \label{fig:JOINT1}
\end{minipage} %
\begin{minipage}{.45\textwidth} %
    \centering
    \includegraphics[scale=0.8]{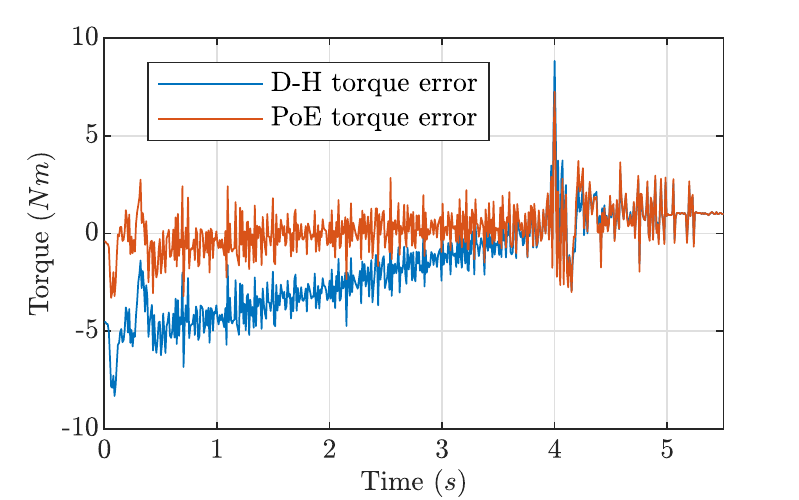}
    \caption{Joint 2 error}
    \label{fig:JOINT2}
\end{minipage} %
\end{figure}

\begin{figure}[h!]
\begin{minipage}{.45\textwidth} %
    \centering
    \includegraphics[scale=0.8]{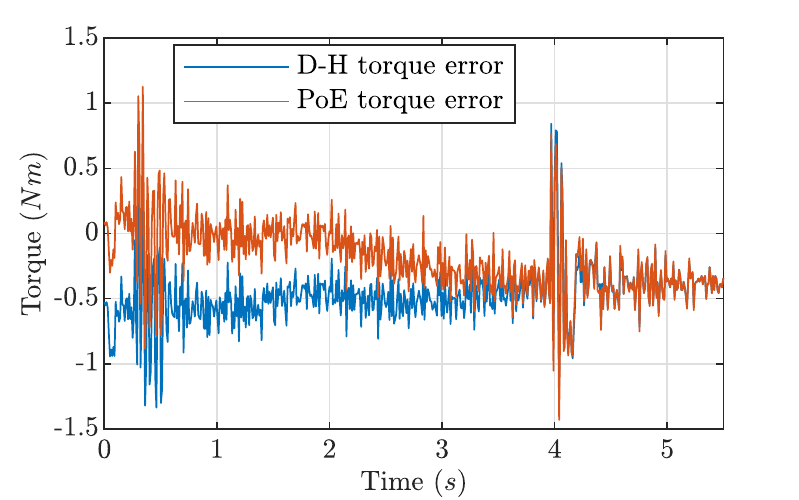}
    \caption{Joint 3 error}
    \label{fig:JOINT3}
\end{minipage} %
\begin{minipage}{.5\textwidth} %
    \centering
    \includegraphics[scale=0.8]{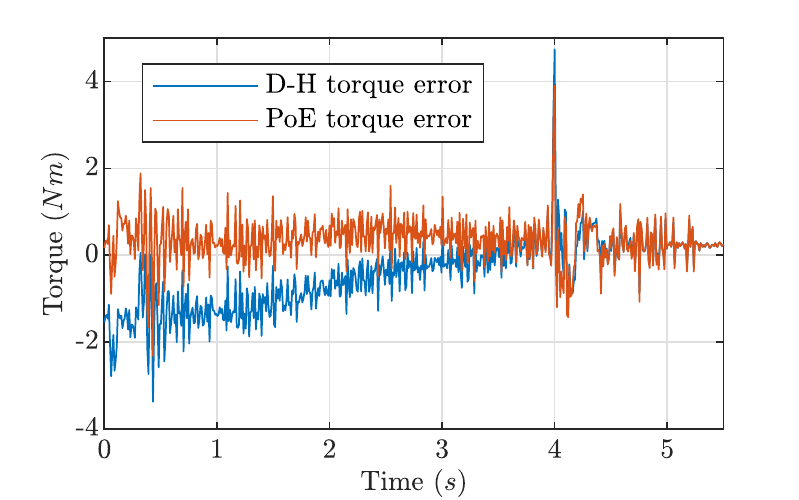}
    \caption{Joint 4 error}
    \label{fig:JOINT4}
\end{minipage} %
\end{figure}

\begin{figure}[h!]
\begin{minipage}{.45\textwidth} %
    \centering
    \includegraphics[scale=0.8]{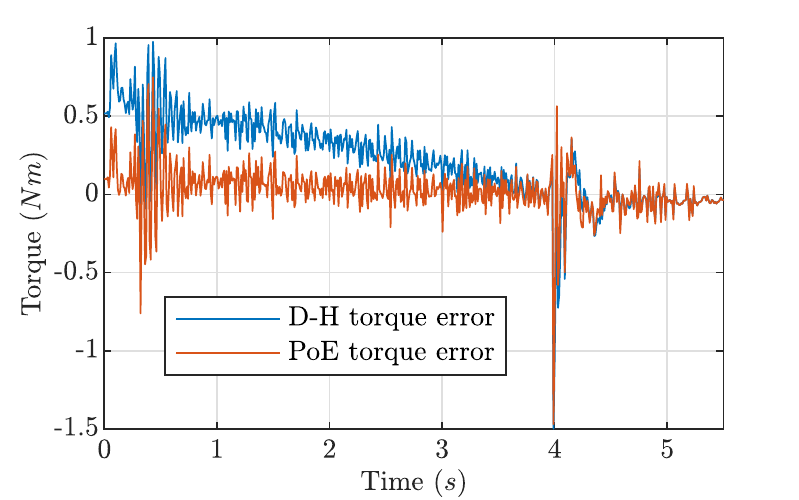}
    \caption{Joint 5 error}
    \label{fig:JOINT5}
\end{minipage} %
\begin{minipage}{.45\textwidth} %
    \centering
    \includegraphics[scale=0.8]{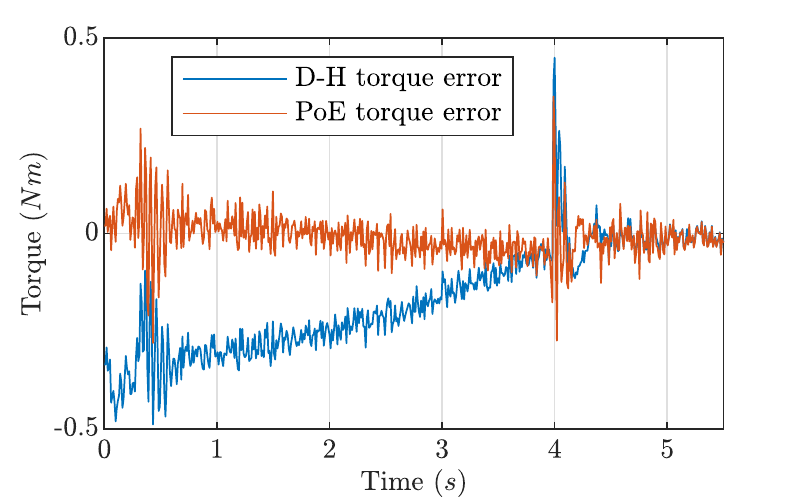}
    \caption{Joint 6 error}
    \label{fig:JOINT6}
\end{minipage} %
\end{figure}

\begin{figure}[h!]
    \centering
    \includegraphics[scale=0.8]{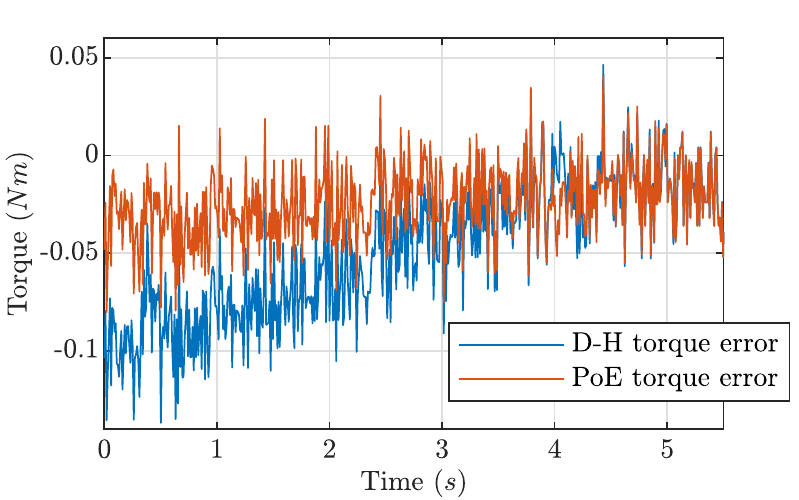}
    \captionof{figure}{Joint 7 error}
    \label{fig:JOINT7}
\end{figure}

The error between the models and the robot can be determined by subtracting the simulation profiles from robot generated torque profiles, \cref{fig:JOINT1} to \cref{fig:JOINT7}. The PoE error (shown in red) is less than D-H parameters error on all of the joints except the first joint, where both methods demonstrated exactly the same degree of accuracy. It is worth to mention that the PoE method exhibits exceptional accuracy in all pitching joints (joints 2,4, and 6), where, in contrast, the D-H parameters method demonstrates high margin of error up to $7\;Nm$ in joint 2 (\cref{fig:JOINT2}).

One of the possible explanations for this error could be the omission of frictional and damping effects of the joints that exist in the real robot that create torques that resist motion, which is not represented in the D-H, PoE, or SimScape\textsuperscript{\textrm{TM}} representations.
The modeling of this phenomena is not in the scope of this primary research, however is being explored to increase model fidelity of future iterations\cite{khan2017review,kermani2007friction,ruderman2014modeling}. The largest error results in joint two, \cref{fig:JOINT2}.

Joint two is the first pitching joint, meaning it is the first joint that resists gravitational effects while moving the subsequent masses. As joint number two is required to move such a large amount of mass, it is expected that this joint will most likely result in the largest torque, which is confirmed by the prediction of the D-H and PoE algorithms (Figures \ref{fig:dh_torq} and \ref{fig:poe_torq}) and shown by the Sawyer torque profile (\cref{fig:exp_torq}).
While profiles appear to be similar in shape, the magnitude, however, is not quite exact. This can be contributed to the lack of damping and friction modelling of the developed algorithms.


%%%%%%%%%%%%%%%%%%%%%%%%%%%%%%%%%%%%%%%%%%%%%%%%%%%%%%%%%%%%%%%%
\section{Platform Dynamics}
To confidently design a controller for the air-bearing platform, the dynamics of simply the robotic arm will not be enough, therefore, the dynamical motion of the air-bearing platform as a system must be described.
The recursive algorithm outlined by Ploen will not work for the entire system as a whole due to the addition of the spherical universal joint, which cannot be properly modeled under that algorithm.
While the spherical air-bearing can be kinematically described as a system of three coincident revolute joints each actuating about a different axis, describing the joints in the recursive algorithm is not as trivial as it sounds, resulting in singularities and yielding incorrect results.
Instead, an analytical dynamic approach was taken to model the system, using the definition of angular momentum\cite{schaub2005analytical} to develop the equations of motion about the spherical air-bearing.

\begin{figure}[h!]
    \centering
    \includegraphics[width=0.6\textwidth]{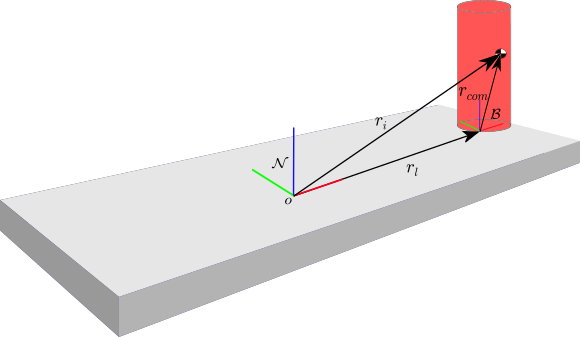}
    \caption{Platform Model Simplification}
    \label{fig:sys}
\end{figure}

To begin with, the first assumption was made that the spherical joint was locked in place, therefore, it cannot rotate. In other words, this means that the torque due to the dynamical motion of the robot and its control box is determined about a fixed point. 
The torque is then determined at the spherical air-bearing, point $o$, due to the motion and weight of each link. To simplify the problem, this process can be conducted using one link at a time first, as seen in \cref{fig:sys}, and then building up to seven links to better resemble the dynamics of the system. 

To correctly represent the inertia of each link about the spherical air-bearing the parallel axis theorem\cite{schaub2005analytical} was used. 
The angular momentum of the $i^{th}$ link about the spherical air-bearing in the body frame can now be determined to be: 

\begin{equation}
    {}^{B}\vec{H}_{o_{i}} = \! {}^{B}I_{com_{i}} \! {}^{B}\vec{\omega}_{i} + m_{i} \left[{}^{B}\vec{r}_{i}^{\; \times} \right] \left[{}^{B}\vec{r}_{i}^{\; \times} \right]^{T} {}^{B}\vec{\omega}_{i}
\end{equation}

where the position of the center of mass $^{B}\vec{r_{i}}$ is: 

\begin{equation}
^{B}\vec{r}_{i} = \! ^{B}\vec{r}_{com_{i}} + \left[R_{N}^{B}\right]^{T} \! ^{N}\vec{r}_{l}
\end{equation}

Using the transport theorem the torque due to rotational motion can now be determined.
Notice that the time derivative of the inertia in the body frame exists, as the parallel axis theorem adds the position vector into the inertia calculation: 

\begin{equation}
  \! ^{N}\dot{\vec{H}}_{o_{i}} =  \frac{d}{dt}\left( \! ^{B}I_{com_{i}} + m_{i} \left[^{B}\vec{r}_{i}^{\; \times} \right] \left[^{B}\vec{r}_{i}^{\; \times} \right]^{T} \right) 
\end{equation}
 
 when expanded it becomes the following:
\begin{equation}
\begin{split}
        \! ^{N}\dot{\vec{H}}_{o_{i}} =  \left({\frac{d}{dt} \! ^{B} \dot{I}_{com_{i}}} + m_{i} \left[^{N}\dot{\vec{r}}_{i}^{\; \times} \right] \left[^{B}\vec{r}_{i}^{\; \times} \right]^{T} + m_{i} \left[^{N}\dot{\vec{r}}_{i}^{\; \times} \right] \left[^{B}\dot{\vec{r}}_{i}^{\; \times} \right]^{T} \right) \! ^{B}\vec{\omega}_{i}+\dots\\
        + \left(^{B} I_{com_{i}} + m_{i} \left[^{B}\vec{r}_{i}^{\; \times} \right] \left[^{B}\vec{r}_{i}^{\; \times} \right]^{T} \right) \!^{B}\dot{\vec{\omega}}_{i} + \!^{N}\vec{\omega}_{i} \times \left(^{B} I_{com_{i}} + m_{i} \left[^{B}\vec{r}_{i}^{\; \times} \right] \left[^{B}\vec{r}_{i}^{\; \times} \right]^{T} \right) \! ^{B}\vec{\omega}_{i}
\end{split}
\end{equation}

where the time derivative of inertia about the center of mass goes to zero $\left({\frac{d}{dt} \! ^{B} \dot{I}_{com_{i}}} = 0\right)$.

\begin{equation}
\begin{split}
    \! ^{N}\dot{\vec{H}}_{o_{i}} = \dot{I}_{o_{i}} \! ^{B}\vec{\omega}_{i} \; + \; \left(^{B} I_{com_{i}} + m_{i} \left[^{B}\vec{r}_{i}^{\; \times} \right] \left[^{B}\vec{r}_{i}^{\; \times} \right]^{T} \right) \!^{B}\dot{\vec{\omega}}_{i} \; +\dots\\ \; \!^{N}\vec{\omega}_{i} \times \left(^{B} I_{com_{i}} + m_{i} \left[^{B}\vec{r}_{i}^{\; \times} \right] \left[^{B}\vec{r}_{i}^{\; \times} \right]^{T} \right) \! ^{B}\vec{\omega}_{i}
\end{split}
\end{equation}

As the rotational torque is the time derivative of angular momentum:
\begin{equation}
    \! ^{N}\vec{T}_{o_{i}} = \! ^{N}\dot{\vec{H}}_{o_{i}}
\end{equation}

 However, this is not the only torque acting on the system, there is also torque due to the weight of each link and the control box that is on the other side of the platform. This torque is calculated using the cross product: 
 
 \begin{equation}
     \! ^{N}\vec{T}_{o_{i}} = \! ^{N}\vec{r}_{i} \times \! ^{N}\vec{F}_{i}
 \end{equation}

Combined gravitational and dynamic torque result in the total torque at the air-bearing that is given below: 

\begin{equation}
    \! ^{N}\tau_{o} = \sum_{1}^{8} \! ^{N}\vec{T}_{o_{i}} =  \sum_{1}^{8} \! ^{N}\dot{\vec{H}}_{o_{i}} + \! ^{N}\vec{r}{i} \times \! ^{N}\vec{F}_{i}
\end{equation}

There are seven links in the robotic arm and the control box can be considered the eighth link although it does not dynamically move; meaning $ \: ^{N}\dot{\vec{H}}_{o_{8}} \rightarrow 0 $.

%%%%%%%%%%%%%%%%%%%%%%%%%%%%%%%%%%%%%%%%%%%%%%%%%%%%%%%%%%%%%%%%%%%%%%
\section{Air-bearing with Robotic Arm Simulation}
\begin{figure}[h!]
    \centering
    \includegraphics[width=0.95\textwidth]{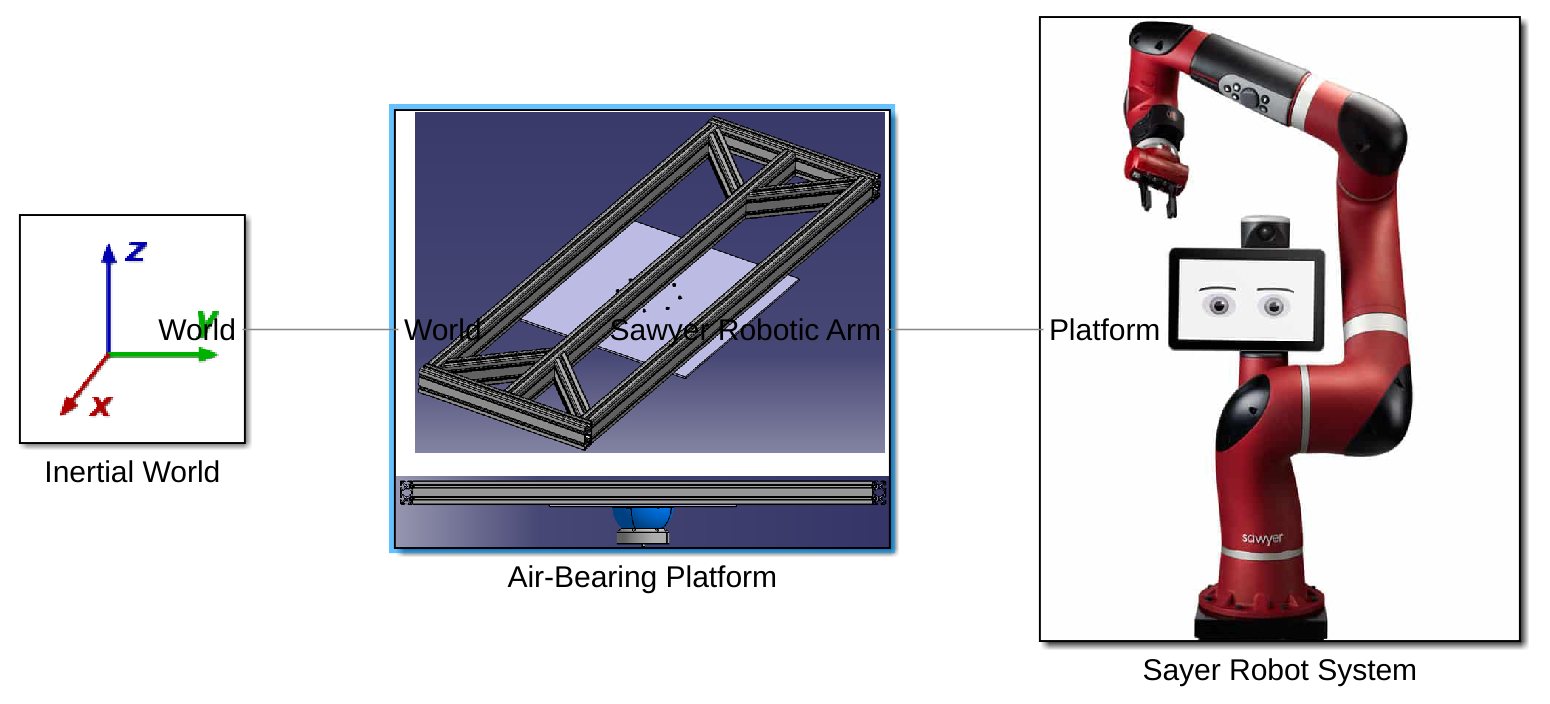}
    \caption{Complete Simulink Model Built in SimScape\textsuperscript{\textrm{TM}}}
    \label{fig:wholesys}
\end{figure}

\begin{figure}[h!]
    \centering
    \includegraphics[width=0.95\textwidth]{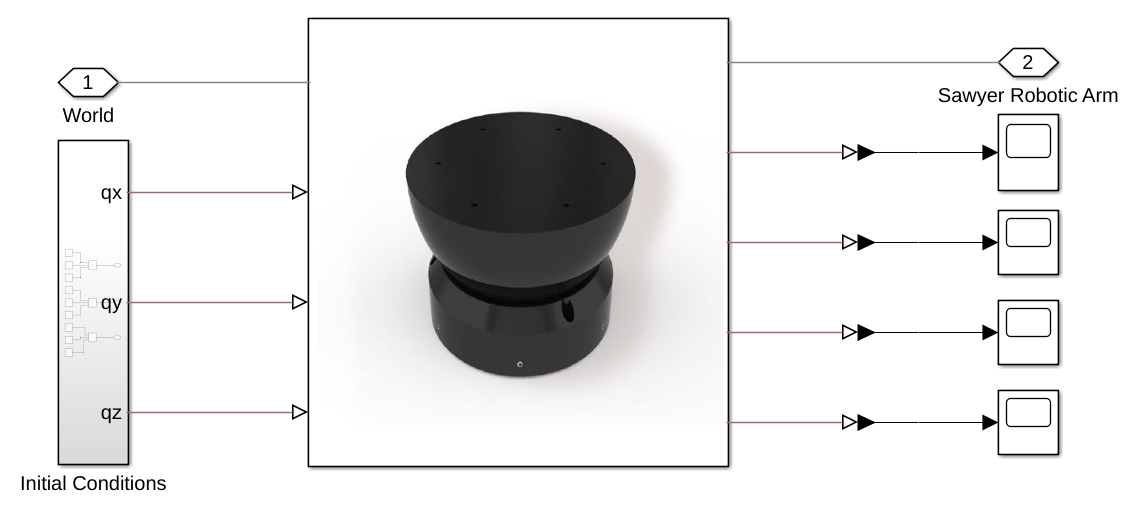}
    \caption{Air-bearing Platform Modeled as a Gimbal}
    \label{fig:Air-Bearing}
\end{figure}

\begin{figure}[h!]
\begin{minipage}{.45\textwidth} %
     \centering
     \includegraphics[scale=.9]{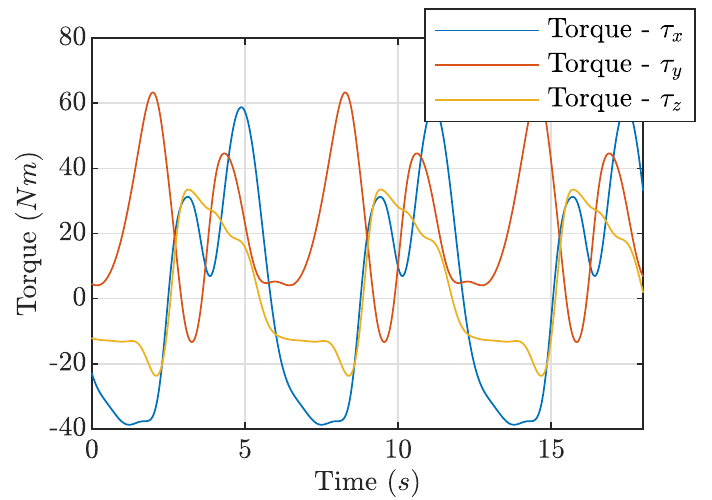}
     \caption{Torque Calculation at the Spherical Air-Bearing}
     \label{fig:Air-Bearing_Torque}
\end{minipage} %
\begin{minipage}{.45\textwidth} %
     \centering
     \includegraphics[scale=.9]{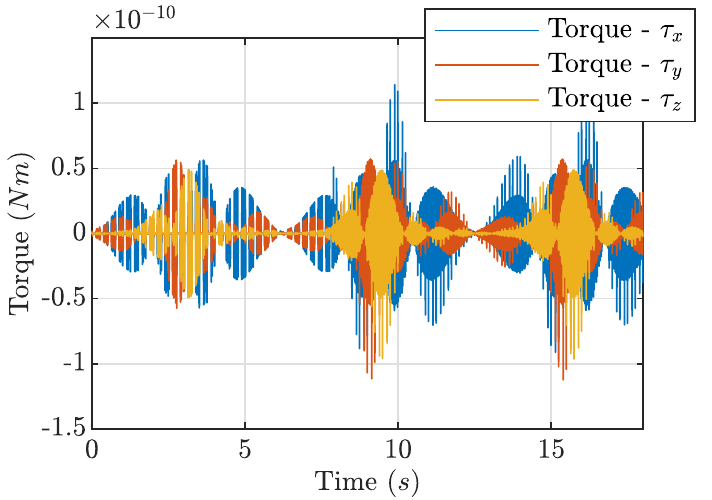}
     \caption{Relative error propagation between theoretical and SimScape\textsuperscript{\textrm{TM}} simulations}
     \label{fig:Air-Bearing_error}
\end{minipage} %
\end{figure}

Using SimScape\textsuperscript{\textrm{TM}}, a model of the system can again be modeled where the air bearing platform is an interface between the robot and the inertial frame, \cref{fig:wholesys}, meaning that there is a displacement in the model from the inertial frame to the robot. The spherical air-bearing is modeled as a gimbal joint (\cref{fig:Air-Bearing}) with the inputs being the trajectory. The trajectory given in \cref{tab:PVA} serves as an input to both theoretical and SimScape\textsuperscript{\textrm{TM}} simulations. Resulting torques about the $x$-, $y$- and $z$-axes are shown in \cref{fig:Air-Bearing_Torque}. The error between the theoretical model and the SimScape\textsuperscript{\textrm{TM}} simulation is negligible as demonstrated in \cref{fig:Air-Bearing_error}. 
This error calculation is generated from the subtraction of the theoretical model results from simulation results.

% \end{minipage} %
% \begin{minipage}{.5\textwidth} %
%     \centering
%     \includegraphics[width=\textwidth]{TSdiff.png}
%     \caption{Relative Error Propagation Between Theoretical and SimScape\textsuperscript{\textrm{TM}} Simulations}
%     \label{fig:Air-Bearing_error}
% \end{minipage}
% \end{figure}

%%%%%%%%%%%%%%%%%%%%%%%%%%%%%%%%%%%%%%%%%%%%%%%%%%%%%%%%%%%%%%%%%
\section{Conclusions}
To ensure the accuracy of analytical dynamic models (D-H parameters and PoE) the methods were compared to a numerical simulation developed in the SimScape\textsuperscript{\textrm{TM}} Multibody environment and experimental results obtained from experiments with Sawyer robotic arm. The comparison showed that the PoE calculates torque profiles in a different way as compared to D-H and SimScape\textsuperscript{\textrm{TM}} approaches. Nevertheless, the PoE approach was more accurate when compared against the experimental results, where the D-H and SimScape\textsuperscript{\textrm{TM}} methods were less accurate. Proving that the PoE approach is ultimately a more accurate approach when it comes to inverse dynamics computation.

The analytical model of the dynamics of the robotic arm on air-bearing table was successfully derived and compared with the SimScape\textsuperscript{\textrm{TM}} simulation, which resulted in good correspondence between the two. The accuracy of the proposed analytical model is planned to be improved by adding damping and frictional effects. The proposed air-bearing test-bed can be used to test relevant space servicing maneuvers and control algorithms for disturbance rejection and attitude control. It is planned to finish manufacturing the air-bearing table to experimentally test these controllers based on the derived analytical model.

%%%%%%%%%%%%%%%%%%%%%%%%%%%%%%%%%%%%%%%%%%%%%%%%%%%%%%%%%%%%%%%%%%%%
\bibliographystyle{AAS_publication}
\bibliography{AAS_References}

\end{document}